\documentclass[a4paper,fleqn]{cas-sc}

\usepackage[authoryear,longnamesfirst]{natbib}
\usepackage{graphicx}
\usepackage{wrapfig}
\usepackage{multirow}
\usepackage{url}
\usepackage{bbm}

\def\tsc#1{\csdef{#1}{\textsc{\lowercase{#1}}\xspace}}
\tsc{WGM}
\tsc{QE}
\tsc{EP}
\tsc{PMS}
\tsc{BEC}
\tsc{DE}

\begin{document}
\let\WriteBookmarks\relax
\def\floatpagepagefraction{1}
\def\textpagefraction{.001}
\shorttitle{Medical Image Analysis (2024)}
\shortauthors{Zehui Liao et~al.}

\title [mode = title]{Unleashing the Potential of Open-set Noisy Samples Against Label Noise for Medical Image Classification}                      
\tnotemark[1]

\tnotetext[1]{
This work was supported in part by the National Natural Science Foundation of China under Grant 62171377, 
the Ningbo Clinical Research Center for Medical Imaging under Grant 2021L003 (Open Project 2022LYKFZD06), 
and the Innovation Foundation for Doctor Dissertation of Northwestern Polytechnical University under Grant CX2023016 and CX2022056.}

\author[1]{Zehui Liao}[type=editor, 
                        auid=000,bioid=1,
                        orcid=0000-0002-8475-5819,
                        ]
\ead{merrical@mail.nwpu.edu.cn}
\affiliation[1]{organization={Northwestern Polytechnical University},
                addressline={1 Dongxiang Road, Chang'an District}, 
                city={Xi'an, Shaanxi},
                postcode={710072}, 
                country={China}}

\author[1]{Shishuai Hu}[
orcid=0000-0002-7314-6647,
]
\ead{sshu@mail.nwpu.edu.cn}

\author[1]{Yanning Zhang}[
orcid=0000-0002-2977-8057,
]
\ead{ynzhang@nwpu.edu.cn}

\author[1,2]{Yong Xia}[
orcid=0000-0001-9273-2847, 
]
\cormark[1]
\ead{yxia@nwpu.edu.cn}

\affiliation[2]{organization={Ningbo Institute of Northwestern Polytechnical University},
                addressline={218 Qingyi Road, Gao'xin District}, 
                city={Ningbo, Zhejiang},
                postcode={315048}, 
                country={China}}

\cortext[cor1]{Corresponding author}

\begin{abstract}
Addressing mixed closed-set and open-set label noise in medical image classification remains a largely unexplored challenge. 
Unlike natural image classification, which often separates and processes closed-set and open-set noisy samples from clean ones, medical image classification contends with high inter-class similarity, complicating the identification of open-set noisy samples. 
Additionally, existing methods often fail to fully utilize open-set noisy samples for label noise mitigation, leading to their exclusion or the application of uniform soft labels.
To address these challenges, we propose the Extended Noise-robust Contrastive and Open-set Feature Augmentation framework for medical image classification tasks. 
This framework incorporates the Extended Noise-robust Supervised Contrastive Loss, which helps differentiate features among both in-distribution and out-of-distribution classes. 
This loss treats open-set noisy samples as an extended class, improving label noise mitigation by weighting contrastive pairs according to label reliability. 
Additionally,  we develop the Open-set Feature Augmentation module that enriches open-set samples at the feature level and then assigns them dynamic class labels, thereby leveraging the model's capacity and reducing overfitting to noisy data.
We evaluated the proposed framework on both a synthetic noisy dataset and a real-world noisy dataset. 
The results indicate the superiority of our framework over four existing methods and the effectiveness of leveraging open-set noisy samples to combat label noise. 
\end{abstract}


\maketitle

\section{Introduction}
\label{sec:intro}

Deep Neural Networks (DNNs) have significantly enhanced the accuracy of medical image classification tasks~\citep{jiang2023review,chen2022recent}. 
The success of these networks largely depends on the accuracy of labeled training data. 
However, the complex nature of medical images and the high level of expertise required for precise annotation often result in noisy labels within clinical datasets~\citep{shi2024survey, karimi2020deep,wang2021annotation,lin2018framework,bai2024improving}. 
Such label noise can lead DNNs to overfit on the training data, which negatively impacts their performance and generalization ability~\citep{zhang2021understanding}.

\begin{figure}[t]
\includegraphics[width=\textwidth]{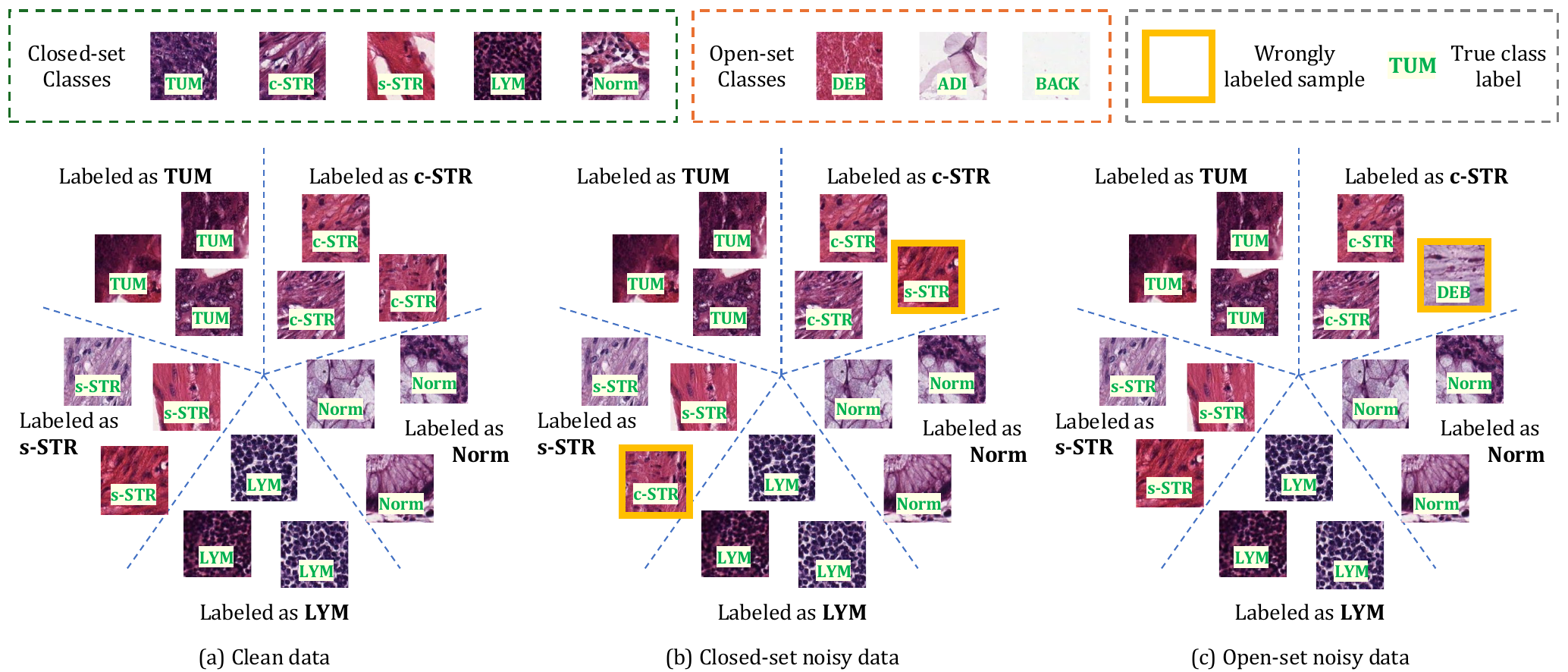}
\caption{
Types of data samples illustrating:
(a) Clean data, where samples are correctly labeled,
(b) Closed-set noisy data, where in-distribution (ID) samples are mislabeled as other known classes, and
(c) Open-set noisy data, where out-of-distribution (OOD) samples are mislabeled as any of the known classes. These samples are sourced from the Kather5k dataset, which comprises eight classes: TUM, s-STR, c-STR, LYM, Norm, DEB, ADI, and Back. To simulate noise, the first five classes are considered ID classes, while the last three are regarded as OOD classes. Each sample is annotated with a \textcolor[RGB]{0,128,0}{green} tag indicating its true class. A \textcolor{orange}{orange} box around an image denotes a mislabeled sample.
}
\label{fig: intro_noise_type}
\end{figure}

\begin{figure}[t]
\centering
\includegraphics[width=\textwidth]{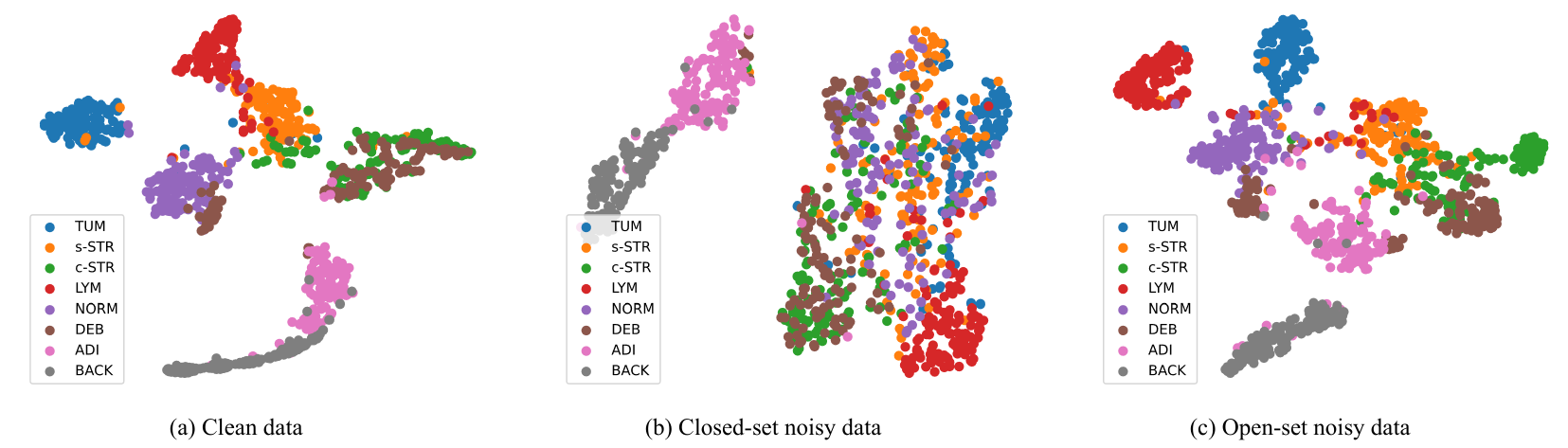}
\caption{
Illustration of feature distributions for both ID and OOD test data from the NoisyKather5k dataset. Sub-figures (a), (b), and (c) depict image features extracted by models trained with clean data, closed-set noisy data, and open-set noisy data, respectively. The closed-set classes consist of TUM, s-STR, c-STR, LYM, and Norm, while the open-set classes comprise DEB, ADI, and Back.
} 
\label{fig: noise_impact}
\end{figure}

Label noise can be categorized into two types: closed-set and open-set. 
As shown in Fig.~\ref{fig: intro_noise_type}, closed-set noise occurs when in-distribution (ID) samples are mislabeled as other known classes, whereas open-set noise involves out-of-distribution (OOD) samples being mislabeled as any known class~\citep{wang2018iterative}. 
The impacts of these noise types are distinct. 
Training a DNN with closed-set noisy data primarily impairs its ability to differentiate between ID categories, but has a relatively minor effect on distinguishing OOD from ID samples (see Fig.~\ref{fig: noise_impact} (b)). 
In contrast, training with open-set noisy data severely hampers the DNN's ability to distinguish between OOD and ID samples, though its performance on ID categories remains relatively intact (see Fig.~\ref{fig: noise_impact} (c)).
For instance, the open-set classes DEB, ADI, and BACK in Fig.~\ref{fig: noise_impact} vary in similarity to the closed-set classes. ADI samples are easily identifiable with a classification model trained on clean or closed-set noisy data but are challenging to classify with a model trained on open-set noisy data.

Despite this, much of the medical research has focused on closed-set noise while neglecting open-set noise~\citep{liu2021co,liao2022learning,ju2022improving,xue2022robust,liu2022nvum,zhang2023trustworthy}. Recent work~\citep{kurian2022improved} acknowledges the presence of both noise types in medical datasets and proposes a method for reweighting noisy samples. However, this approach treats both noise types uniformly, ignoring their differences.

In natural image classification, prevalent methods differentiate between closed-set or open-set noisy samples and clean ones with distinct strategies~\citep{sun2020crssc,sachdeva2021EDM,yao2021jo,sun2022PNP,albert2023PLS}. These methods leverage the unique characteristics of each noise type: high training losses are typically associated with closed-set noise because DNNs tend to fit these noisy labels during later training stages~\citep{liu2020early}, while high predictive uncertainty is linked to open-set noise due to the DNNs' lack of familiarity with OOD classes~\citep{kendall2017uncertainties,wimmer2023quantifying}. Strategies for managing closed-set noise include pseudo-label generation~\citep{albert2023PLS,sun2022PNP,sachdeva2021EDM,sun2020crssc,yao2021jo}, sample reweighting~\citep{albert2023PLS}, and noise-robust loss functions~\citep{liao2022ILDR}. Conversely, methods for open-set noise often involve discarding samples~\citep{albert2023PLS,sachdeva2021EDM,sun2020crssc} or assigning nearly uniform soft labels~\citep{sun2022PNP,yao2021jo}. 
While these techniques improve performance in natural image classification, they face two significant challenges in medical imaging: (1) the high inter-class similarity among medical images complicates the identification of open-set noisy samples; and (2) the potential advantages of leveraging these identified open-set samples to further mitigate label noise have yet to be thoroughly explored.

\begin{figure}[t]
\centering
\includegraphics[width=1.0\textwidth]{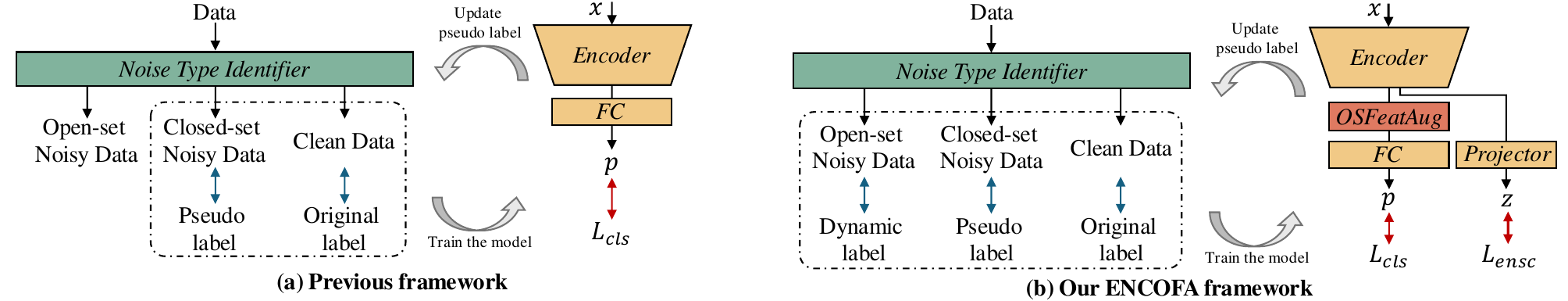}
\caption{
Comparison between (a) a representative previous framework and (b) our ENCOFA framework designed to address mixed open-set and closed-set label noise. The previous framework distinguishes among clean samples, closed-set noisy samples, and open-set noisy samples. It trains the model using clean samples and closed-set noisy samples. In contrast, our ENCOFA framework enhances the model by introducing an extended noise-robust supervised contrastive loss $L_{ensc}$ alongside the classification loss $L_{cls}$. This approach significantly improves the identification accuracy of open-set noisy samples. Furthermore, ENCOFA utilizes detected open-set noisy samples to enrich them through the OSFeatAug module. 
The enriched open-set features are assigned dynamic labels, effectively consuming the model’s additional capacity and thereby preventing overfitting.
} 
\label{fig: intro_diff_ours}
\end{figure}

In this paper, we propose the \textbf{E}xtended \textbf{N}oise-robust \textbf{C}ontrastive and \textbf{O}pen-set \textbf{F}eature \textbf{A}ugmentation (\textbf{ENCOFA}) framework to tackle the challenges of mixed closed-set and open-set label noise in medical image classification. 
As shown in Fig.~\ref{fig: intro_diff_ours}, ENCOFA categorizes training samples into three categories: clean, closed-set noisy, and open-set noisy. It then trains a classification network, consisting of an encoder and a fully connected layer, using observed labels, generated pseudo labels, and dynamic class labels for each category, respectively.
To address closed-set noise, we employ Supervised Contrastive Learning (SCL)\citep{khosla2020supervised}, which creates contrastive pairs to enhance intra-class cohesion and inter-class separation. Although effective, SCL's inherent limitations in handling OOD classes and susceptibility to label noise necessitate further refinement. Therefore, we design the Extended Noise-robust Supervised Contrastive (ENSC) Loss, which extends the class space to include detected open-set samples and adjusts contrastive pair weights based on label reliability.
For open-set noise, we consume the extra capacity of DNNs by incorporating dynamically labeled open-set samples into the training process~\citep{wei2021open}. To this end, we construct the Open-set Feature Augmentation (OSFeatAug) module, which augments open-set noisy samples at the feature level and optimizes the augmentation probability to balance the quantity of augmented samples. This approach helps mitigate model overfitting to noise and enhances robustness.
Our experiments on both the synthetic NoisyKather5K~\citep{kather2016multi} and real-world noisy Kaggle DR~\citep{galdran2022test} datasets validate the superior performance of the ENCOFA framework compared to existing methods and highlight the effectiveness of its components.

The main contributions of this work are four-fold:
\begin{itemize}
    \item We propose the ENCOFA framework to address the mixed closed-set and open-set label noise issue, which has rarely been explored in medical image classification.
    \item To enhance the identification accuracy of open-set noisy samples in medical image classification tasks characterized by high inter-class similarity, we design the ENSC loss that tolerates label noise and separates features of different classes, covering both closed-set and open-set classes.
    \item To leverage the detected open-set samples, we construct the OSFeatAug module to enrich them at the feature level and assign them dynamic labels, effectively consuming the extra capacity of the model and preventing overfitting to noisy data.
    \item Experimental results on both synthetic and real-world noisy datasets demonstrate that our ENCOFA framework outperforms existing methods in handling the mixed closed-set and open-set label noise in medical image classification.
\end{itemize}

\section{Related Work}
\label{sec:related_work}

\subsection{Learning with Open-set and Closed-set Label Noise}
Most existing approaches assume closed-set label noise, where mislabeled samples are still within the known class set~\citep{han2018co,wang2019symmetric,li2019dividemix,liu2020early,zhou2021asymmetric,karim2022unicon,huang2023twin,wu2024time,wang2024tackling}. This assumption can be overly restrictive since, in many cases, the true classes of mislabeled samples may fall outside the known class set, leading to an open-set label noise problem~\citep{sukhbaatar2015training,wang2018iterative}.

To address mixed open-set and closed-set label noise, several methods have been proposed, including modeling noise transition matrices~\citep{sukhbaatar2015training,xia2022extended,yao2023LCCN}. However, real-world label noise patterns are often complex and variable, making this task challenging. Alternatively, many methods focus on identifying open-set noisy samples and closed-set noisy samples and processing different types of noisy samples separately~\citep{sachdeva2021EDM,sun2022PNP,yao2021jo,sun2020crssc,albert2023PLS}. For noise type identification, the EvidentialMix (EDM) method~\citep{sachdeva2021EDM} uses subjective logic loss to distinguish between sample types based on statistical differences in loss values. The Joint Sample Selection and Model Regularization based on Consistency (Jo-SRC) method~\citep{yao2021jo} and the Probabilistic Noise Prediction (PNP) method~\citep{sun2022PNP} link noisy samples to discrepancies between labels and predictions of weakly augmented views, while open-set noise is identified through high divergence between predictions from two augmented views. The Certainty-based Reusable Sample Selection and Correction (CRSSC) approach~\citep{sun2020crssc} and the Pseudo Loss Selection (PLS) method~\citep{albert2023PLS} differentiate samples based on training loss values and prediction certainty. 

Although these methods handle closed-set noisy samples through pseudo-label generation~\citep{albert2023PLS,sun2022PNP,sachdeva2021EDM,sun2020crssc} and sample reweighting~\citep{albert2023PLS}, they often discard open-set noisy samples~\citep{albert2023PLS,sachdeva2021EDM} or assign them smooth soft labels~\citep{sun2022PNP}. Furthermore, these techniques struggle with high inter-class similarity in medical images, leading to suboptimal identification accuracy for open-set noisy samples. To address these issues, our ENCOFA framework introduces an Extended Noise-robust Supervised Contrastive Loss to improve the distinction of open-set samples and an Open-set Feature Augmentation module to enrich detected open-set noisy samples which are assigned dynamic labels for mitigating overfitting to label noise.

\subsection{Label Noise-robust Medical Image Classification}
To enhance the performance and generalization of deep learning algorithms in medical image classification tasks, it is essential to address the issue of label noise, which frequently arises due to low-quality annotations. 
Various methods have been developed to combat label noise, including modeling noise transition matrices~\citep{tanno2019learning}, noise-robust regularization~\citep{pornvoraphat2023real}, sample re-weighting~\citep{ju2022improving,kurian2022improved,xue2019robust}, sample selection~\citep{xue2022robust,gong2022less}, negative learning with attention mechanism~\citep{liao2022learning}, and label correction techniques~\citep{ai2023url,qiu2023hierarchical,zhu2021hard,liu2021co}. 
However, most of these methods focus primarily on closed-set label noise and do not fully address open-set label noise, which can also occur in medical image classification tasks. 
Although some researchers have acknowledged open-set label noise, they often overlook the differences between open-set and closed-set noisy samples, resulting in a unified sample reweighting framework to address noisy samples~\citep{kurian2022improved}. In contrast, our ENCOFA framework specifically targets the unique characteristics of both open-set and closed-set noisy samples, employing tailored strategies for each type to effectively mitigate the impact of label noise on model performance.

\begin{figure*}[t]
\includegraphics[width=\textwidth]{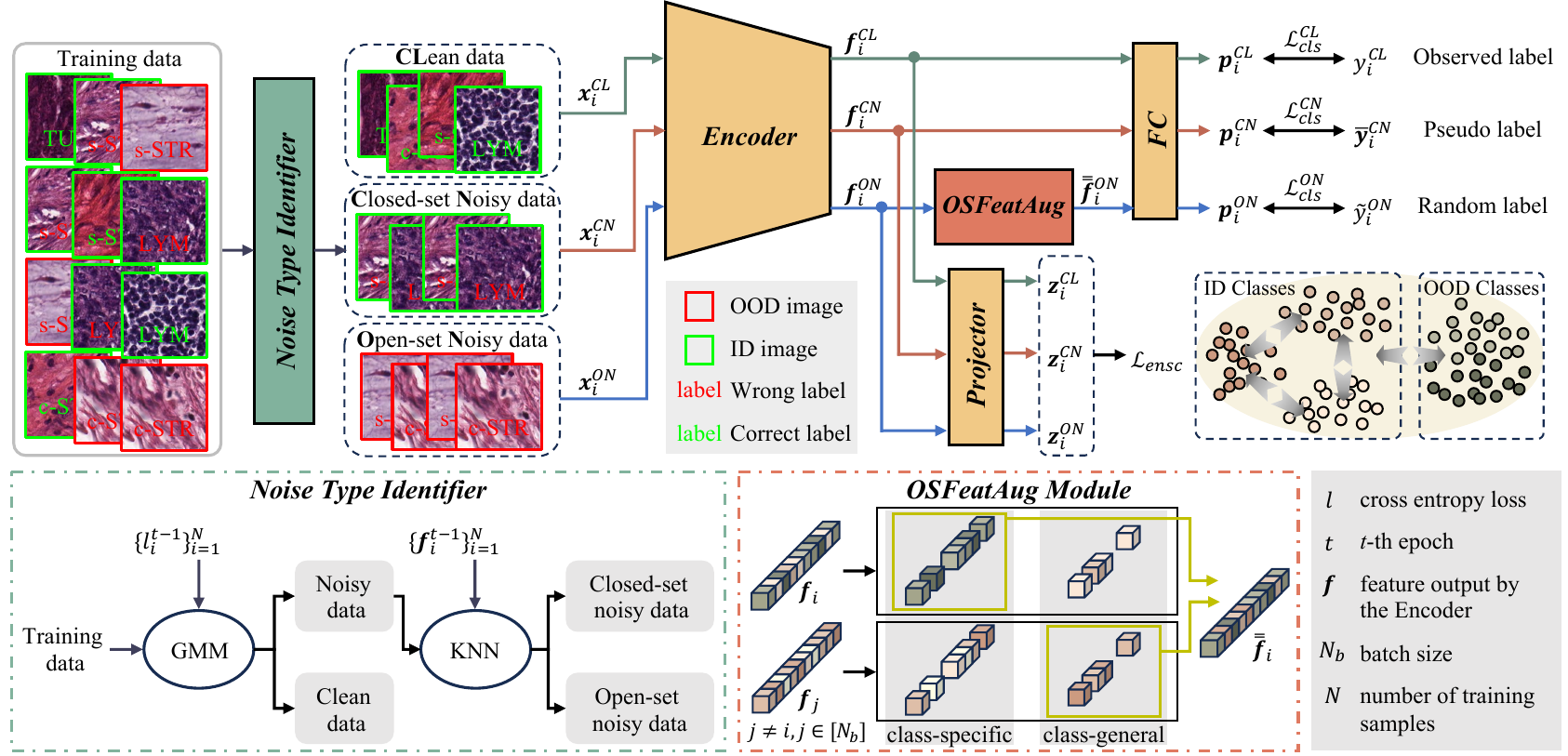}
\caption{
Illustration of the ENCOFA framework.
Samples are classified into clean, closed-set noisy, or open-set noisy categories using a noise type identifier, and then processed through an encoder and a FC layer.
Open-set noisy sample features are augmented using the OSFeatAug module.
Classification losses for clean, closed-set noisy, and open-set noisy samples are computed based on their observed, pseudo, or random labels, respectively.
Meanwhile, the output features from the projector are utilized to calculate the ENSC loss.
} 
\label{fig: overview}
\end{figure*}

\section{Method}
\label{sec:method}
\subsection{Problem Formalization and Method Overview}
We address a $K$-class medical image classification task using a noisy dataset $\textit{D} = \left \{ \left ( \textit{\textbf{x}}_i, y_i \right )  \right \} ^ {N} _{i=1}$, comprising $N$ image-label pairs. 
Here, $\textit{\textbf{x}}_i \in \mathbb{R}^{C \times H \times W}$ denotes the $i$-th image with dimensions $H \times W$ and $C$ channels, while $y_i \in [K] = \{ 1,2,\ldots,K \}$ represents its observed label. 
Our objective is to develop a precise medical image classification model by training on this noisy dataset $\textit{D}$, which includes clean samples, closed-set noisy samples, and open-set noisy samples.

The proposed ENCOFA framework, depicted in Fig.~\ref{fig: overview}, comprises a noise type identifier, an encoder coupled with a fully connected (FC) layer, an OSFeatAug module, and a projector. 
The framework begins with a warm-up phase over several epochs, followed by the main training phase.
During each epoch of the training phase, two primary steps are performed: first, the noise type identifier classifies samples into clean, closed-set noisy, and open-set noisy categories; second, all samples are processed by the encoder and FC layer, with open-set noisy samples undergoing additional enhancement through the OSFeatAug module. Supervision is applied to clean, closed-set noisy, and open-set noisy samples using observed labels, generated pseudo labels, and dynamic labels, respectively. Additionally, the ENSC loss is computed using the output features from the projector. We will now detail each component of the framework.

\subsection{Classification Backbone}
The ResNet architecture~\citep{he2016deep} serves as the basis of our classification backbone, incorporating an encoder $\mathcal{F}_E(\Theta_E)$ and a FC layer $\mathcal{F}_{FC}(\Theta_{FC})$, where $\Theta_E$ and $\Theta_{FC}$ denote their respective parameters. 
The encoder contains a single convolutional block and four residual blocks followed by an average pooling layer. 
For an input image $\textit{\textbf{x}}_i$, the encoder extracts the feature vector 
\begin{equation}
    \textbf{\textit{f}}_i = \mathcal{F}_E(\textit{\textbf{x}}_i; \Theta_E), 
\end{equation}
which is then processed by the FC layer and a softmax function $\mathcal{S}$, yielding the probabilistic output 
\begin{equation}
    \textbf{\textit{p}}_i = \mathcal{S}(\mathcal{F}_{FC}(\textbf{\textit{f}}_i; \Theta_{FC})).
\end{equation}

\subsubsection{Noise Type Identifier}
The noise type identifier operates in two steps. 
First, the training data is partitioned into clean and noisy groups using a Gaussian Mixture Model (GMM). 
Next, a K-Nearest Neighbors (KNN)-based OOD detection method is applied to identify open-set samples within the noisy group.

\noindent \textbf{GMM.}
Observations suggest that DNNs often fit clean data before noisy data, implying that noisy samples typically exhibit higher loss values than clean samples during early training stages~\citep{liu2020early}.
Therefore, at the beginning of each epoch, we categorize all training samples into clean and noisy groups based on their loss values $\{l_i^{t-1}\}_{i=1}^N$ from the previous epoch using a two-component GMM. 
Here, $l_i^{t-1}$ represents the cross-entropy (CE) loss value of sample $\textbf{\textit{x}}_i$, calculated between the probabilistic output $\textbf{\textit{p}}_i$ and its observed label $y_i$ in the $(t-1)$-th epoch.

Assuming that $\{l_i^{t-1}\}_{i=1}^N$ follows a mixture of two Gaussian distributions, we estimate the parameters of the GMM using the Expectation-Maximization (EM) algorithm. We then compute the posterior probability $p(\mathcal{G} \mid l_i^{t-1})$ that the loss value $l_i^{t-1}$ belongs to the Gaussian component $\mathcal{G}$ with a lower mean, which represents the distribution of smaller losses from clean samples. To accurately identify clean samples, we select a threshold value $\gamma_{CL} \in (0.5, 1.0)$ and classify samples with a loss value whose posterior probability is greater than $\gamma_{CL}$ as clean; otherwise, they are categorized as noisy.
Let $I=[N]$ and $I^{CL}$ denote the index sets for all training samples and detected clean samples.

\noindent \textbf{KNN.}
A KNN-based OOD detection method~\citep{sun2022KNN,zhang2023openood} is then employed to distinguish open-set samples from noisy samples, based on the assumption that OOD samples are relatively distant from ID data in the feature space. This method calculates the OOD score for each sample and classifies those with OOD scores below a predefined threshold $\gamma_{OOD}$ as open-set.
The decision function is given by:
\begin{equation}
    G(\textbf{\textit{f}}, k) = \mathbbm{1}\{- \left \| \mathcal{N}(\textbf{\textit{f}}) - \mathcal{N}(\textbf{\textit{f}}_{(k)}) \right \| _2 \le \gamma_{OOD}\}
\end{equation}
where $\textbf{\textit{f}} \in \{\textbf{\textit{f}}_i^{t-1}\}_{i \in [N]\setminus I^{CL}}$ represents the input feature vector, and $\textbf{\textit{f}}_{(k)} \in \{\textbf{\textit{f}}_i^{t-1}\}_{i \in I^{CL}}$ is its $k$-th nearest ID neighbor, with $k$ is set to 200 in this study.
The function $\mathcal{N}$ denotes normalization, and $\textbf{\textit{f}}_i^{t-1}$ is the feature of $\textit{\textbf{x}}_i$ extracted by the encoder in the previous epoch.
The Euclidean distance between the normalized $\textbf{\textit{f}}$ and $\textbf{\textit{f}}_{(k)}$ is calculated, and the negative distance is compared to $\gamma_{OOD}$. 
The indicator function $\mathbbm{1}[A]$ represents whether the condition $A$ is met. 
The threshold $\gamma_{OOD}$, independent of OOD data, is typically chosen to ensure a high classification accuracy for ID data ($e.g.$, 95\%).
Let $I^{CN}$ and $ I^{ON}$ denote the index sets for detected closed-set noisy samples and open-set noisy samples.

\subsubsection{Classification Loss}
For each clean sample $(\textit{\textbf{x}}_i^{CL}, y_i^{CL})$, we aim to minimize the CE loss between its prediction $\textbf{\textit{p}}_i^{CL}$ and observed label $y_i^{CL}$. For each closed-set noisy sample $(\textit{\textbf{x}}_i^{CN}, y_i^{CN})$, the loss is minimized between the prediction and the generated soft pseudo label $\bar{\textbf{\textit{y}}}_i^{CN} \in [0.0,1.0]^K$. For each open-set noisy sample $(\textit{\textbf{x}}_i^{ON}, y_i^{ON})$, the loss is minimized between the prediction and a dynamic label  $\tilde{y}_i^{ON}$ which is randomly selected from $[K]$.
The classification loss is computed as follows:
\begin{equation}
    \mathcal{L}_{cls} = \mathcal{L}_{cls}^{CL} + \mathcal{L}_{cls}^{CN} + \mathcal{L}_{cls}^{ON} = \sum_{i\in I^{CL}}\mathcal{L}(y_i^{CL}, \textbf{\textit{p}}_i^{CL}) + \sum_{i\in I^{CN}}\omega_i^{CN}\mathcal{L}(\bar{\textbf{\textit{y}}}_i^{CN}, \textbf{\textit{p}}_i^{CN}) + \sum_{i\in I^{ON}}\mathcal{L}(\tilde{y}_i^{ON}, \textbf{\textit{p}}_i^{ON})
\end{equation}
where $\mathcal{L}$ denotes the CE loss.

\noindent \textbf{Pseudo Label Generation.}
For each closed-set noisy sample $\textit{\textbf{x}}_i^{CN}$ (omitting $CN$ for simplicity), we feed its two weakly augmented views $\textit{\textbf{x}}_i^{v1}$ and $\textit{\textbf{x}}_i^{v2}$ into the backbone.
Data augmentation strategies include random horizontal flips and rotations ranging from -10 to 10 degrees.
The predictions $\textbf{\textit{p}}_i^{v1}$ and $\textbf{\textit{p}}_i^{v2}$ from these views are used to create the pseudo label:
\begin{equation}
    \bar{\textbf{\textit{y}}}_i = \left \| ((\textbf{\textit{p}}_i^{v1}+\textbf{\textit{p}}_i^{v2}) \div 2)^{2} \right \|_{1},
\end{equation}
where $\left \| \cdot  \right \| _{1}$ denotes L1 normalization, ensuring that the class probabilities in the soft label sum to 1.0.
Since pseudo labels may not always be accurate, a second GMM is used to fit pseudo-label losses. The probability that a pseudo-label loss falls into the lower-mean component (indicating higher reliability) is used as the weight for the loss term $\mathcal{L}(\bar{\textbf{\textit{y}}}_i^{CN}, \textbf{\textit{p}}_i^{CN})$, denoted as $\omega_i^{CN}$.

\subsection{Extended Noise-robust Supervised Contrastive Loss}
The ENSC loss is one of the key components of the proposed ENCOFA framework, differentiating our ENCOFA from existing methods by improving the poor identification accuracy of open-set noisy samples via supervised contrastive learning for medical image classification tasks characterized by high inter-class similarity.
Accurate labels are crucial for supervised contrastive learning to prevent the undue reduction of distances between features from different classes. 
Therefore, correcting incorrect labels in the training set is essential before computing the ENSC loss.
For closed-set noisy samples, we utilize the class with the highest probability in their pseudo labels for supervised contrastive learning.
Notably, open-set noisy samples are categorized into an additional $(K+1)$-th class, enabling the ENSC loss not only to differentiate between ID classes but also to segregate ID from OOD classes.
The refined labels are expressed as $y'_i \in [K+1]$, for $i\in I$. 
Furthermore, the reliability of $y’_i$ informs the assignment of a weight factor $\omega_i$ for each sample $\textit{\textbf{x}}_i$: a weight of 1.0 is assigned to both clean and open-set noisy samples, while the weight for closed-set noisy samples, as previously discussed, enhances the ENSC loss’s robustness to noise, thereby improving its accuracy. The ENSC loss is computed as follows:
\begin{equation}
    \mathcal{L}_{ensc} = \sum_{i\in I}{\frac{-1}{|P(i)|}}\sum_{p \in P(i)}{\log \frac{\omega_i \omega_p \cdot exp(\textbf{\textit{z}}_i \cdot \textbf{\textit{z}}_p / \tau)}{\sum_{a \in A(i)}{\omega_i \omega_a \cdot exp(\textbf{\textit{z}}_i \cdot \textbf{\textit{z}}_a / \tau)}}},
\end{equation}
where $\textbf{\textit{z}}_i$ is the output feature of the projector given $\textbf{\textit{f}}_i$, 
the projector contains an FC layer followed by a normalization function, 
$A_{i}\equiv I \setminus {i}$ is the index set of all training samples without $i$-th sample, 
$P(i) \equiv \{p \in A(i):y'_p=y'_i\}$ is the index set of training samples in the same class as feature $\textbf{\textit{z}}_i$, 
and $\tau \in \mathbb{R}^{+}$ is a scalar temperature parameter that is set to 0.2 in this study.
As the value of ENSC loss decreases during training, features of the same category will be pulled closer together, while features of different categories will be pushed further apart, including open-set classes.

\subsection{OSFeatAug Module}
The OSFeatAug module is the other key component of the proposed ENCOFA framework.
Rather than disregarding detected open-set noisy samples as in existing methods, our OSFeatAug
module is designed to augment the features of open-set noisy samples, thereby expanding the effective number of open-set noisy samples, which are assigned dynamic labels to consume the model’s excess capacity and reduce overfitting to noisy training data.
Concretely, we start by using gradient-based methods to identify which channels of the feature $\textbf{\textit{f}}_i \in \mathbb{R}^{d}$ extracted by the encoder are class-specific and which are class-general. 
Specifically, we record the gradient $\{\textbf{\textit{g}}_i^{t-1}\}_{i=1}^N$ of the predictions with respect to the feature map of the last convolutional layer for all training samples, where $\textbf{\textit{g}}_i^{t-1} \in \mathbb{R}^{d \times H' \times W'}$. 
These gradients are then global-average-pooled over the width and height dimensions to obtain the channel importance weights~\citep{selvaraju2017grad}, denoted as $\bar{\textbf{\textit{g}}}_i^{t-1} \in \mathbb{R}^{d}$.
The average importance weights of channels across all training samples are calculated as:
\begin{equation}
    \hat{\textbf{\textit{g}}}^{t-1} = \frac{1}{N} \sum_{N}^{i=1} \bar{\textbf{\textit{g}}}_i^{t-1},
\end{equation}
and then $\hat{\textbf{\textit{g}}}^{t-1}$ is min-max-normalized and compared with a threshold $\gamma_{gen} \in (0.0, 1.0)$. 
Channels with average importance weights below $\gamma_{gen}$ are treated as class-general.

Once the class-specific and class-general channels are identified, each open-set noisy sample $\textbf{\textit{f}}_i^{ON}$ (omitting $ON$ for simplicity) fed into the OSFeatAug module retains its class-specific channels while its class-general channels are substituted with the class-general channels from another feature $\textbf{\textit{f}}_j, j\ne i \wedge j\in [N_b]$ (randomly selected from the same batch). 
Here, $N_b$ denotes the batch size.
This process ensures that the essential characteristics of the open-set noisy samples remain intact while enhancing variability and diversity in the feature representations, effectively increasing the number of distinct open-set noisy samples encountered by the model during training. 
The input open-set noisy feature is augmented with a probability $\gamma_p \in [0.0, 1.0]$. 
The augmented feature, denoted by $\bar{\bar{f}}_i$, is then used for predictions by the FC layer.

\subsection{Loss Function}
The loss function of our ENCOFA framework comprises two components: the classification loss $\mathcal{L}_{cls}$ and the extended noise-robust supervised contrastive loss $\mathcal{L}_{ensc}$. 
These components are combined using a weighting factor $\lambda$, shown as follows
\begin{equation}
    \mathcal{L} = \mathcal{L}_{cls} + \lambda * \mathcal{L}_{ensc}.
\end{equation}

\section{Datasets}
\label{sec:dataset}
\noindent\textbf{Synthetic Noisy Dataset.} 
The Kather-5k dataset~\citep{kather2016multi} consists of 5,000 class-balanced pathological image patches, each measuring $150 \times 150$ pixels, and categorized into eight classes: tumor epithelium (TUM), simple stroma (s-STR), complex stroma (c-STR), immune cell conglomerates (LYM), normal colon mucosa (NORM), debris (DEB), adipose tissue (ADI), and background tissue (BACK). 
The first five classes are designated as closed-set, while the remaining three are open-set~\citep{galdran2022test}\footnote{This classification follows advice from pathologists to simulate a scenario where only clinically relevant tissue regions are labeled.}.
The closed-set samples are split into training, validation, and test sets with ratios of 70\%, 10\%, and 20\%, respectively. 
The open-set samples are used to introduce open-set label noise into the training and validation sets. Following strategies from previous studies~\citep{cheng2022instance,li2023disc}, we apply instance-dependent label noise, altering labels for closed-set noise and images for open-set noise, with sample-specific probabilities. In NoisyKather5k, the noise rate is denoted as $\alpha$, with $\beta$ representing the proportion of open-set noise and $(1-\beta)$ corresponding to closed-set noise.

\noindent\textbf{Real-World Noisy Dataset.}
The Diabetic Retinopathy (DR) dataset~\citep{Emma2015KaggleDR} contains 88,702 fundus images from 44,351 patients, with each patient contributing one photograph per eye.
These fundus images are classified into five categories of DR: Normal, mild Non-proliferative DR (NPDRI), moderate Non-proliferative DR (NPDRII), severe Non-proliferative DR (NPDRIII), and proliferative DR (PDR).
Label noise, estimated at 30-40\%, originates from observer variability within DR classes ($i.e.$, closed-set noise) and mislabeling of other retinal diseases as DR ($i.e.$, open-set noise). 
\citet{ju2022improving} re-labeled 57,213 images from this dataset, with 917 of these images confirmed by multiple experts and used as the gold standard test set. The remaining 56,296 images, with their original DR labels, constitute the noisy training dataset. Due to an overrepresentation of normal class samples (42,185), we selected 8,406 Normal cases to create a final training set of 22,517 samples, with 10\% allocated for validation.
This study addresses both binary classification (Normal vs. DR) and multi-class classification tasks.

\begin{table}[t]
\caption{
Hyper-parameter settings for our ENCOFA framework on the NoisyKather5k dataset with varying noise rates ($\alpha \in [0.2, 0.4]$, $\beta \in [0.25, 0.50, 0.75]$).
}
\label{tab: HyperparameterKatheer5k}
\centering
\renewcommand\arraystretch{1.25}
\setlength\tabcolsep{6.0pt}
\begin{tabular}{cc|c}
\hline \hline
\multicolumn{2}{c|}{Noise rate}                                     & Hyper-parameters' setting \\ \hline \hline
\multicolumn{1}{c|}{\multirow{3}{*}{$\alpha$=0.2}} & $\beta$=0.25   & $\gamma_{CL}$=0.98, $\gamma_{OOD}$=0.97, $\gamma_{gen}$=0.1, $\gamma_p$=0.7, $\lambda$=1.0 \\ \cline{2-3} 
\multicolumn{1}{c|}{}                              & $\beta$=0.50   & $\gamma_{CL}$=0.98, $\gamma_{OOD}$=0.96, $\gamma_{gen}$=0.2, $\gamma_p$=0.7, $\lambda$=0.5 \\ \cline{2-3} 
\multicolumn{1}{c|}{}                              & $\beta$=0.75   & $\gamma_{CL}$=0.98, $\gamma_{OOD}$=0.97, $\gamma_{gen}$=0.1, $\gamma_p$=0.7, $\lambda$=1.0 \\ \hline
\multicolumn{1}{c|}{\multirow{3}{*}{$\alpha$=0.4}} & $\beta$=0.25   & $\gamma_{CL}$=0.98, $\gamma_{OOD}$=0.96, $\gamma_{gen}$=0.1, $\gamma_p$=0.7, $\lambda$=1.0 \\ \cline{2-3} 
\multicolumn{1}{c|}{}                              & $\beta$=0.50   & $\gamma_{CL}$=0.98, $\gamma_{OOD}$=0.96, $\gamma_{gen}$=0.1, $\gamma_p$=0.9, $\lambda$=1.5 \\ \cline{2-3} 
\multicolumn{1}{c|}{}                              & $\beta$=0.75   & $\gamma_{CL}$=0.98, $\gamma_{OOD}$=0.96, $\gamma_{gen}$=0.2, $\gamma_p$=0.7, $\lambda$=0.5  \\ \hline \hline
\end{tabular}
\end{table}

\section{Experiments and Results}

\subsection{Implementation Details}
For the NoisyKather-5k dataset, we utilized ResNet18 as the backbone architecture, initializing ENCOFA with a 1-epoch warm-up and training for a total of 200 epochs. 
For the DR dataset, ResNet34 was employed as the backbone, with ENCOFA similarly warmed up for 10 epochs and trained for 100 epochs. All images were resized to $224 \times 224$ pixels. 
The initial learning rate $lr$ was set to $1 \times 10^{-2}$ for NoisyKather-5k and $3 \times 10^{-4}$ for DR, following a polynomial decay policy defined as $lr = lr_{0} \times \left(1 - t/T \right)^{0.9}$, where $t$ represents the current epoch and $T$ denotes the total number of epochs. We set the batch size to 128 and used the Adam optimizer~\citep{Kingma2015AdamAM} with a weight decay of $1 \times 10^{-4}$ for all experiments.
The hyperparameters for ENCOFA were set as follows: $\gamma_{CL}$=0.75, $\gamma_{OOD}$=0.98, $\gamma_{gen}$=0.1, $\gamma_p$=0.3, $\lambda$=1.5 for the DR dataset. For the NoisyKather-5k dataset, the hyperparameters are detailed in Table~\ref{tab: HyperparameterKatheer5k}.
All results are reported based on three independent runs, with both the mean and standard deviation provided.

The primary evaluation metric is the classification accuracy on the test set, denoted by Acc$^{test}$. For performance analysis during training, we also report the classification accuracy on the validation set Acc$^{val}$, the noise type classification accuracy Acc$_{type}^{train}$, and the F1-score F1$_{ON}^{train}$ and precision Pre$_{ON}^{train}$ for identifying open-set noisy samples.

\begin{table*}[t]
\caption{
Test accuracy (\%, mean$\pm$standard deviation) of our ENCOFA and competing methods on the NoisyKather5k dataset, with noise levels $\alpha \in \{0.2, 0.4\}$ and corruption rates $\beta \in \{0.25, 0.50, 0.75\}$.
The results are highlighted in \textbf{bold} for the best performance in each column.
The number of trainable model parameters (M) and GPU memory cost (GB) for all competing methods are reported.
}
\label{tab: Comparison_NoisyKather5k}
\centering
\footnotesize
\renewcommand\arraystretch{1.25}
\setlength\tabcolsep{3pt}
\begin{tabular}{l|ccc|ccc|c|c}
\hline \hline
\multirow{2}{*}{Method} & \multicolumn{3}{c|}{$\alpha$=0.2}          & \multicolumn{3}{c|}{$\alpha$=0.4}             & \multirow{2}{*}{Params} & \multirow{2}{*}{GPU cost}  \\ \cline{2-7}
                  & \multicolumn{1}{c|}{$\beta$=0.25} & \multicolumn{1}{c|}{$\beta$=0.50} & $\beta$=0.75 & \multicolumn{1}{c|}{$\beta$=0.25} & \multicolumn{1}{c|}{$\beta$=0.50} & $\beta$=0.75 &             &         \\ \hline \hline
CE                        & \multicolumn{1}{c|}{78.72$\pm$0.45} & \multicolumn{1}{c|}{81.44$\pm$1.90} & 84.43$\pm$0.72 & \multicolumn{1}{c|}{63.20$\pm$1.07} & \multicolumn{1}{c|}{70.08$\pm$1.36} & 76.69$\pm$1.7  & 11.18 & 2.61 \\ \hline
ILDR~\citep{liao2022ILDR}  & \multicolumn{1}{c|}{90.72$\pm$0.13} & \multicolumn{1}{c|}{89.39$\pm$0.98} & 90.93$\pm$0.95 & \multicolumn{1}{c|}{83.73$\pm$3.38} & \multicolumn{1}{c|}{86.61$\pm$1.55} & 88.11$\pm$0.75 & 11.18 & 2.61  \\ \hline
EDM~\citep{sachdeva2021EDM}& \multicolumn{1}{c|}{91.95$\pm$0.72} & \multicolumn{1}{c|}{91.09$\pm$1.06} & 90.51$\pm$1.88 & \multicolumn{1}{c|}{83.95$\pm$2.42} & \multicolumn{1}{c|}{88.05$\pm$0.30} & 89.55$\pm$0.72 & 22.36 & 6.60 \\ \hline
PNP~\citep{sun2022PNP}     & \multicolumn{1}{c|}{91.15$\pm$0.20} & \multicolumn{1}{c|}{91.25$\pm$0.98} & 91.68$\pm$0.99 & \multicolumn{1}{c|}{77.65$\pm$6.99} & \multicolumn{1}{c|}{86.67$\pm$2.16} & 89.33$\pm$1.31 & 11.71 & 7.25 \\ \hline
PLS~\citep{albert2023PLS}  & \multicolumn{1}{c|}{93.60$\pm$0.82} & \multicolumn{1}{c|}{93.76$\pm$0.39} & 93.23$\pm$1.00 & \multicolumn{1}{c|}{87.47$\pm$1.07} & \multicolumn{1}{c|}{90.13$\pm$0.87} & 90.61$\pm$0.72 & 11.24 & 5.68  \\ \hline
Ours                      & \multicolumn{1}{c|}{\textbf{94.36}$\pm$0.13} & \multicolumn{1}{c|}{\textbf{94.46}$\pm$0.42} & \textbf{94.65}$\pm$0.53 & \multicolumn{1}{c|}{\textbf{91.73}$\pm$0.62} & \multicolumn{1}{c|}{\textbf{91.57}$\pm$1.27} & \textbf{92.10}$\pm$0.66 & 11.24 & 5.41  \\ \hline \hline
\end{tabular}
\end{table*}

\begin{table}[t]
\caption{
Noise type classification accuracy and F1-score of open-set noisy sample identification (\%, mean$\pm$standard deviation) of ENCOFA and three recent competing methods on the NoisyKather5k dataset ($\alpha$=0.4, $\beta$=0.25).
The best result is highlighted in \textbf{bold}.
}
\label{tab: comparison_noise_type_cls}
\centering
\renewcommand\arraystretch{1.25}
\setlength\tabcolsep{22.0pt}
\footnotesize
\begin{tabular}{l|c|c}
\hline \hline
Method                      & Acc$_{type}^{train}$  & F1$_{ON}^{train}$   \\ \hline \hline
EDM~\citep{sachdeva2021EDM}  & 68.37$\pm$2.26        & 41.43$\pm$11.61       \\ \hline
PNP~\citep{sun2022PNP}       & 74.32$\pm$6.81        & 0.00$\pm$0.00       \\ \hline
PLS~\citep{albert2023PLS}    & 76.82$\pm$2.70        & 0.19$\pm$0.01       \\ \hline
Ours                        & \textbf{91.94}$\pm$1.18        & \textbf{77.85}$\pm$4.25        \\ \hline \hline
\end{tabular}
\end{table}

\subsection{Comparative Experiments}
We evaluated our ENCOFA framework against a baseline method and four recent approaches on both the NoisyKather-5k and DR datasets. The baseline method involves a ResNet trained on the noisy dataset using the standard CE loss. Four competing methods include one method for handling closed-set label noise (\textit{i.e.}, Instance-dependent Label Distribution Regularization (ILDR)\citep{liao2022ILDR}) and three methods addressing mixed closed-set and open-set label noise (\textit{i.e.}, EvidentialMix (EDM)\citep{sachdeva2021EDM}, Probabilistic Noise Prediction (PNP)\citep{sun2022PNP}, and Pseudo Loss Selection (PLS)\citep{albert2023PLS}).
Our ENCOFA framework distinguishes itself from these competing methods by notably improving the identification accuracy of open-set noisy samples through the introduction of the ENSC loss. 
Additionally, it effectively leverages identified open-set noisy samples by enriching them using the OSFeatAug module and utilizing them to reduce the impact of label noise.

\noindent \textbf{Results on Synthetic Noisy Dataset.}
Table~\ref{tab: Comparison_NoisyKather5k} presents the results on the NoisyKather5k dataset, with noise rates set at 0.2 or 0.4 and open-set noise proportions ranging from 0.25 to 0.75. The results indicate that ENCOFA outperforms all competing methods. Notably, in the most challenging scenario ($\alpha$=0.4, $\beta$=0.25), ENCOFA achieves a significant performance improvement over its competitors, with an accuracy of 91.73\% compared to 87.47\% (p-value=0.0078$<$0.05).

Additionally, since the true noisy types of training samples are known in the synthetic noisy dataset, we compare our ENCOFA framework with other competing methods based on noise type classification accuracy and the F1-score of open-set noisy sample detection. 
As shown in Table~\ref{tab: comparison_noise_type_cls}, the proposed ENCOFA framework significantly outperforms other competitors in terms of Acc$_{type}^{train}$ and F1$_{ON}^{train}$. 
This improvement is due to the ENSC loss in ENCOFA, which enhances inter-class separation and intra-class cohesion. 
Note that ILDR is not included in this comparison because it does not distinguish between closed-set and open-set noisy data during training.

\begin{table}[t]
\caption{
Test Accuracy (\%, mean $\pm$ standard deviation) of our ENCOFA framework and competing methods on the Kaggle DR dataset, evaluated for both binary and five-class classification tasks. The best result in each column is highlighted in \textbf{bold}.
}
\label{tab: Comparison_KaggleDRplus}
\centering
\footnotesize
\renewcommand\arraystretch{1.25}
\setlength\tabcolsep{20.0pt}
\begin{tabular}{l|c|c}
\hline \hline
Method                      & Five-class task   & Two-class task     \\ \hline \hline
CE                          & 46.57$\pm$0.79    & 63.79$\pm$1.37     \\ \hline
ILDR~\citep{liao2022ILDR}    & 54.67$\pm$0.72    & 70.01$\pm$0.39     \\ \hline
EDM~\citep{sachdeva2021EDM}  & 57.86$\pm$0.82    & 71.66$\pm$0.54     \\ \hline
PNP~\citep{sun2022PNP}       & 57.92$\pm$0.49    & 68.54$\pm$0.49     \\ \hline
PLS~\citep{albert2023PLS}    & 57.78$\pm$1.17    & 70.88$\pm$0.31     \\ \hline 
Ours                        & \textbf{60.18}$\pm$0.78    & \textbf{72.85}$\pm$0.45     \\ \hline \hline
\end{tabular}
\end{table}

\noindent \textbf{Results on Real-World Noisy Dataset.} 
We also evaluated our ENCOFA, the baseline, and four competing methods on the real-world Kaggle DR dataset. Table~\ref{tab: Comparison_KaggleDRplus} shows the test accuracies for both binary and five-class classification tasks. The results reveal that ENCOFA surpasses the best competitor by 2.26\% (p-value=0.0189$<$0.05) on the binary classification task and by 1.19\% (p-value=0.0444$<$0.05) on the five-class classification task.
The results suggest that the proposed ENCOFA can help the trained classifier combat real-world label noise, which is instance-dependent and more challenging than synthetic label noise.

\begin{table}[t]
\caption{
Test accuracy and Validation Accuracy (\%, mean $\pm$ standard deviation) of ENCOFA and its two variants, as well as $\mathcal{L}_{cls}$ and its two variants, evaluated on the NoisyKather5k dataset ($\alpha$=0.4, $\beta$=0.25).
The best result is highlighted in \textbf{bold}.
}
\label{tab: Ablation_NoisyKather5k}
\centering
\footnotesize
\renewcommand\arraystretch{1.25}
\setlength\tabcolsep{9.5pt}
\begin{tabular}{ccc|c|c|c|c}
\hline \hline
\multicolumn{3}{c|}{$\mathcal{L}_{cls}$}                                                        & \multirow{2}{*}{$\mathcal{L}_{ensc}$} & \multirow{2}{*}{OSFeatAug} & \multirow{2}{*}{Acc$^{val}$} & \multirow{2}{*}{Acc$^{test}$} \\ \cline{1-3}
\multicolumn{1}{c|}{$\mathcal{L}_{cls}^{CL}$}  & \multicolumn{1}{c|}{$\mathcal{L}_{cls}^{CN}$}  & $\mathcal{L}_{cls}^{ON}$  &                     &                  &                              &                               \\ \hline \hline
\multicolumn{1}{c|}{\checkmark} & \multicolumn{1}{c|}{}                          &                           &                                       &               & 40.89$\pm$2.57               & 63.20$\pm$1.07                \\ \hline
\multicolumn{1}{c|}{\checkmark} & \multicolumn{1}{c|}{\checkmark} &                           &                                       &                              & 56.45$\pm$3.70               & 87.20$\pm$1.16                \\ \hline
\multicolumn{1}{c|}{\checkmark} & \multicolumn{1}{c|}{\checkmark} & \checkmark &                                       &                                             & 53.00$\pm$1.65               & 86.19$\pm$1.00                \\ \hline \hline
\multicolumn{3}{c|}{\checkmark}                                                                              &                                       &               & 53.00$\pm$1.65               & 86.19$\pm$1.00                \\ \hline
\multicolumn{3}{c|}{\checkmark}                                                                              & \checkmark             &                              & 57.33$\pm$1.96               & 90.93$\pm$1.96                              \\ \hline
\multicolumn{3}{c|}{\checkmark}                                                                              & \checkmark                            & \checkmark    & \textbf{57.71}$\pm$1.10               & \textbf{91.73}$\pm$0.62                 \\ \hline \hline
\end{tabular}
\end{table}

\begin{table}[t]
\caption{
Impact of identified CL, CN, and ON samples on test Accuracy, validation accuracy, noise type classification accuracy, and precision of open-set noisy sample identification (\%, mean$\pm$standard deviation) using $\mathcal{L}_{cls}$ with $\mathcal{L}_{ensc}$ trained on the NoisyKather5k dataset ($\alpha$=0.4, $\beta$=0.25). 
The best result is highlighted in \textbf{bold}.
}
\label{tab: Discuss_Noisy_SCL}
\centering
\footnotesize
\renewcommand\arraystretch{1.25}
\setlength\tabcolsep{9pt}
\begin{tabular}{c|c|c|c|c|c|c}
\hline \hline
CL    & ON      & CN                           & Acc$^{val}$       & Acc$^{test}$          & Acc$_{type}^{train}$   & Pre$_{ON}^{train}$ \\ \hline \hline
              &                 &              & 53.00$\pm$1.65    & 86.19$\pm$1.00        & 87.17$\pm$1.89         & 69.69$\pm$9.12  \\ \hline 
\checkmark    &                 &              & 57.11$\pm$1.26    & 89.65$\pm$0.20        & 88.41$\pm$2.22         & 69.14$\pm$8.99  \\ \hline
\checkmark    & \checkmark      &              & 57.18$\pm$1.75    & 89.81$\pm$0.20        & 89.29$\pm$1.68         & 85.07$\pm$5.67  \\ \hline
\checkmark    & \checkmark      & \checkmark   & \textbf{57.33}$\pm$1.96    & \textbf{90.93}$\pm$1.96        & \textbf{91.93}$\pm$2.30         & \textbf{85.84}$\pm$5.03  \\ \hline
\hline
\end{tabular}
\end{table}

\subsection{Ablation Analysis}
\label{sec: ablation}
We conducted ablation studies on the NoisyKather5k dataset with mixed label noise ($\alpha$=0.4, $\beta$=0.25) to assess the contribution of each component within our ENCOFA framework. 
Table~\ref{tab: Ablation_NoisyKather5k} summarizes the performance of ENCOFA and its variants.
The notation `$\mathcal{L}_{cls}$+$\mathcal{L}_{ensc}$+OSFeatAug' denotes the complete ENCOFA framework; 
`$\mathcal{L}_{cls}$+$\mathcal{L}_{ensc}$' represents ENCOFA without the OSFeatAug module; 
and `$\mathcal{L}_{cls}$' refers to the model trained solely with the classification loss. 
The results indicate that ENCOFA's performance is significantly enhanced by integrating either $\mathcal{L}_{ensc}$ or the OSFeatAug module.

\begin{figure*}[t]
\centering
\includegraphics[width=\textwidth]{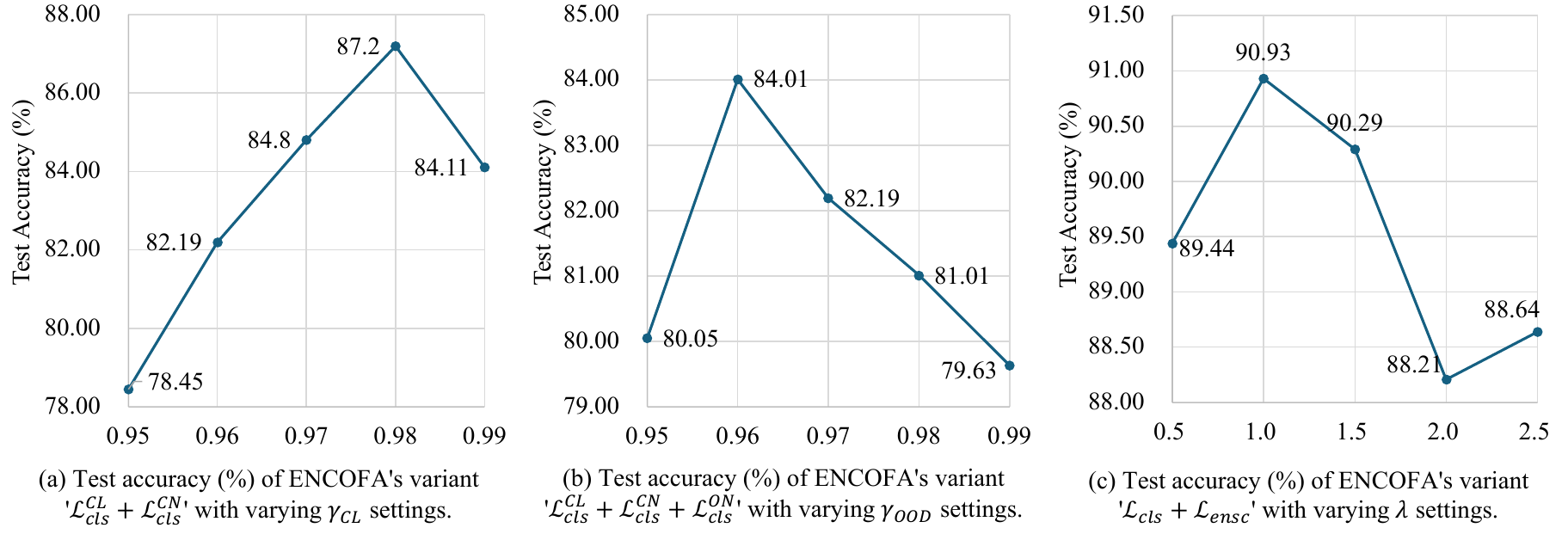}
\caption{
Performance comparison of ENCOFA variants with respect to (a) $\gamma_{CL}$, (b) $\gamma_{OOD}$, and (c) $\lambda$, evaluated on the NoisyKather5k dataset ($\alpha$=0.4, $\beta$=0.25).
The variant `$\mathcal{L}_{cls}^{CL}+\mathcal{L}_{cls}^{CN}$' identifies noisy samples within the training data and utilizes both clean and noisy data to train the classification backbone, supervised by observed and pseudo labels, respectively.
The variant `$\mathcal{L}_{cls}^{CL}+\mathcal{L}_{cls}^{CN}+\mathcal{L}_{cls}^{ON}$' detects closed-set and open-set noisy samples across all training data, with these two types of noisy data supervised by pseudo and dynamic labels, respectively.
The variant `$\mathcal{L}_{cls}+\mathcal{L}_{ensc}$' integrates the extended noise-robust supervised contrastive loss, building upon `$\mathcal{L}_{cls}^{CL}+\mathcal{L}_{cls}^{CN}+\mathcal{L}_{cls}^{ON}$'.
}
\label{fig: hyper_parameter}
\end{figure*}

\begin{wrapfigure}{r}{0.45\textwidth} 
\centering
\includegraphics[width=0.45\textwidth]{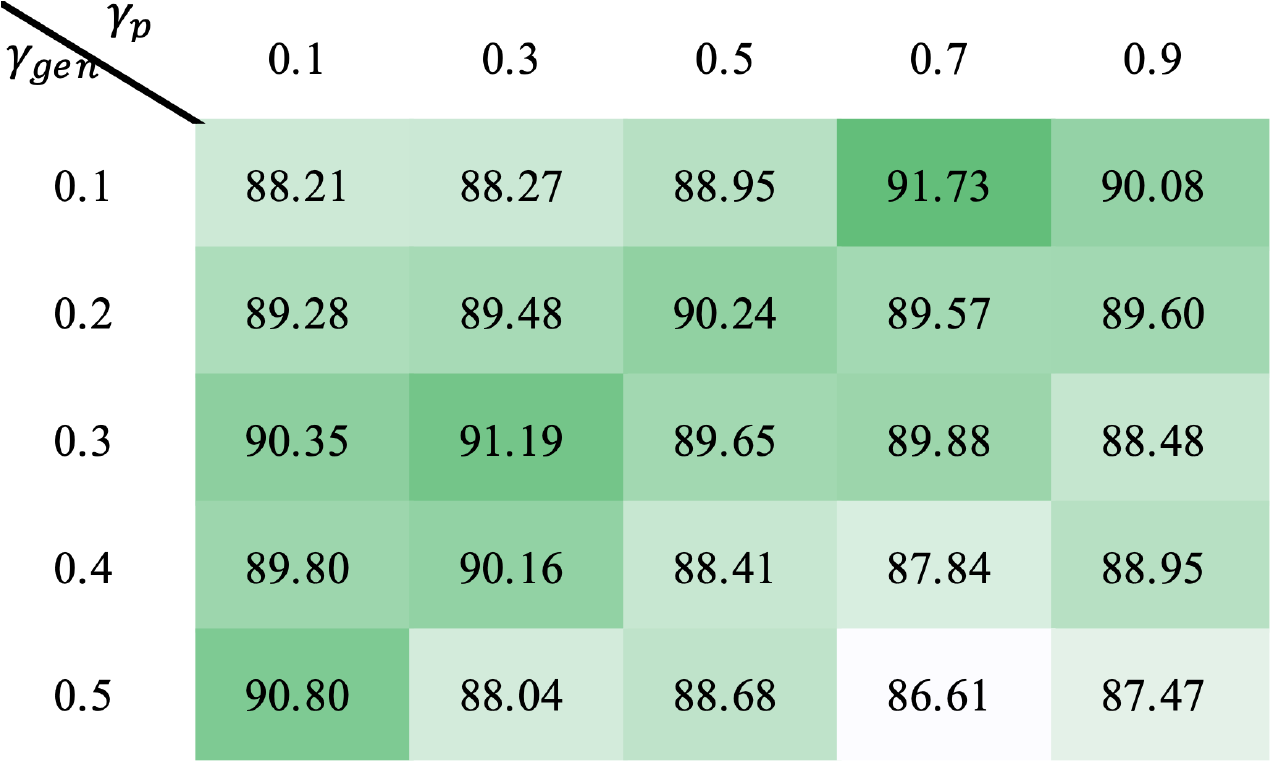}
\caption{
Test accuracy (\%) of the proposed ENCOFA framework with varying settings of $\gamma_{gen}$ and $\gamma_p$ on the NoisyKather5k dataset ($\alpha$=0.4, $\beta$=0.25).
}
\label{fig: gamma_gen_gamma_p}
\end{wrapfigure}

Additionally, we evaluated the performance of `$\mathcal{L}_{cls}$' and its variants: `$\mathcal{L}_{cls}^{CL}$', which computes the CE loss for all samples using their observed labels; 
`$\mathcal{L}_{cls}^{CL}$+$\mathcal{L}_{cls}^{CN}$', which uses observed and pseudo labels for clean and noisy samples, respectively; 
and `$\mathcal{L}_{cls}^{CL}$+$\mathcal{L}_{cls}^{CN}$+$\mathcal{L}_{cls}^{ON}$', which employs observed, pseudo, and dynamic labels for clean, closed-set noisy, and open-set noisy samples, respectively. 
The results in Table~\ref{tab: Ablation_NoisyKather5k} show that incorporating $\mathcal{L}_{cls}^{CN}$ to $\mathcal{L}_{cls}^{CL}$ brings a significant performance gain.
However, when $\mathcal{L}_{cls}^{ON}$ is further introduced, the performance is slightly reduced.
This can be attributed to inaccuracies in distinguishing noise types. 
Misclassification of some closed-set samples as open-set results in these samples being incorrectly supervised with dynamic labels, thereby reducing performance.
The contribution of $\mathcal{L}_{cls}^{ON}$ becomes evident after improving noise type classification accuracy by introducing the ENSC loss $\mathcal{L}_{ensc}$.

\section{Discussion}
\subsection{Analysis of $\mathcal{L}_{ensc}$}
We evaluated the impact of clean, closed-set noisy, and open-set noisy samples on optimizing $\mathcal{L}_{ensc}$, focusing on test accuracy Acc$^{test}$ and the precision of open-set noisy sample identification Pre$_{ON}^{train}$.
Note that ground truth noise types for training samples are not available during training.
Table~\ref{tab: Discuss_Noisy_SCL} details the performance of $\mathcal{L}_{cls}$ with the complete $\mathcal{L}_{ensc}$. 
The results indicate that minimizing $\mathcal{L}_{ensc}$ using only clean samples significantly improves test accuracy, while the precision Pre$_{ON}^{train}$ of detecting open-set noisy samples remained relatively unchanged.
After including the detected open-set noisy samples, $\mathcal{L}_{ensc}$ enhances inter-class separation across all closed-set and open-set classes, significantly improving the precision of open-set noisy sample identification.
Furthermore, integrating closed-set noisy samples with pseudo labels and pairwise weights further improves both test accuracy and noise type classification accuracy.

\begin{table}[t]
\caption{
Test accuracy, validation accuracy, noise type classification accuracy, and precision of open-set noisy sample identification (\%, mean$\pm$standard deviation) achieved by our ENCOFA framework, with and without employing clustering on identified open-set noisy samples, evaluated on the NoisyKather5k dataset ($\alpha$=0.4, $\beta$=0.25).
}
\label{tab: clustering}
\centering
\footnotesize
\renewcommand\arraystretch{1.25}
\setlength\tabcolsep{9pt}
\begin{tabular}{l|c|c|c|c}
\hline \hline
$\mathcal{L}_{ensc}$ & Acc$^{val}$    & Acc$^{test}$ & Acc$_{type}^{train}$ & Pre$_{ON}^{train}$ \\ \hline \hline
w.o. clustering  & 57.71$\pm$1.10 & 91.73$\pm$0.62  & 91.94$\pm$1.18   & 85.95$\pm$4.09  \\ \hline
w. clustering    & 56.67$\pm$1.89 & 90.45$\pm$0.33  & 89.95$\pm$1.59   & 82.61$\pm$3.69  \\ \hline \hline
\end{tabular}
\end{table}

\subsection{Open-set Noisy Samples Clustering}
Our ENCOFA framework detects open-set noisy samples and treats them as a unified OOD class.
To investigate the effect of clustering identified open-set noisy samples during training, we tested ENCOFA with clustering applied post-open-set-noisy-sample-detection on the NoisyKather5k dataset ($\alpha$=0.4, $\beta$=0.25). We used the K-Means clustering algorithm with the number of clusters set to 3, corresponding to the known number of OOD classes. However, as shown in Table~\ref{tab: clustering}, clustering did not enhance performance. This lack of improvement is attributed to the fact that clustering detected OOD samples did not affect either the detection or utilization of OOD samples.

\subsection{Hyper-parameter Selection}
The proposed ENCOFA framework requires tuning five hyper-parameters, including $\gamma_{CL}$ for clean sample identification, $\gamma_{OOD}$ for open-set noisy sample identification, $\lambda$ for weighting the loss term $\mathcal{L}_{ensc}$, $\gamma_{gen}$ for class-general channel selection, and $\gamma_{p}$ for adjusting the probability of performing feature augmentation for open-set noisy samples.
During the development of this framework, we adjusted the hyper-parameters of the new components while keeping the parameters of the previous models fixed.
As described in Section~\ref{sec: ablation}, ENCOFA and its variants are evaluated as follows: (a) $\mathcal{L}_{cls}^{CL}$, (b) $\mathcal{L}_{cls}^{CL}+\mathcal{L}_{cls}^{CN}$, (c) $\mathcal{L}_{cls}^{CL}+\mathcal{L}_{cls}^{CN}+\mathcal{L}_{cls}^{ON}$, (d) $\mathcal{L}_{cls}+\mathcal{L}_{ensc}$, and (e) $\mathcal{L}_{cls}+\mathcal{L}_{ensc}$+OSFeatAug (ENCOFA).
From model $\mathcal{L}_{cls}^{CL}$ to model $\mathcal{L}_{cls}^{CL}+\mathcal{L}_{cls}^{CN}$, we introduced a detection step to distinguish noisy data from clean data, focusing solely on adjusting the hyperparameter $\gamma_{CL}$ (see Fig.~\ref{fig: hyper_parameter}(a)). 
Next, from model $\mathcal{L}_{cls}^{CL}+\mathcal{L}_{cls}^{CN}$ to model $\mathcal{L}_{cls}^{CL}+\mathcal{L}_{cls}^{CN}+\mathcal{L}_{cls}^{ON}$, we incorporated another detection step to identify open-set noisy samples from noisy samples, requiring only the adjustment of the newly introduced hyperparameter $\gamma_{OOD}$ (see Fig.~\ref{fig: hyper_parameter}(b)), while keeping the value of $\gamma_{CL}$ unchanged.
Subsequently, from model $\mathcal{L}_{cls}^{CL}+\mathcal{L}_{cls}^{CN}+\mathcal{L}_{cls}^{ON}$ to model $\mathcal{L}_{cls}+\mathcal{L}_{ensc}$, we incorporated the ENSC loss and only adjusted its weight factor $\lambda$ (see Fig.~\ref{fig: hyper_parameter}(c)).
Finally, from model $\mathcal{L}_{cls}+\mathcal{L}_{ensc}$ to our ENCOFA, we introduced the OSFeatAug module and selected the value of $\gamma_{gen}$ and $\gamma_p$ using the grid search strategy (see Fig.~\ref{fig: gamma_gen_gamma_p}).
In this study, we tuned all hyper-parameters of our ENCOFA on the NoisyKather5k and DR datasets with the following search ranges: $\gamma_{CL} \in \{0.65, 0.75, 0.85, 0.95, 0.96, 0.97, 0.98, 0.99\}$, $\gamma_{OOD} \in \{0.95, 0.96, 0.97, 0.98, 0.99\}$, $\lambda \in \{0.5, 1.0, 1.5, 2.0, 2.5\}$, $\gamma_{gen} \in \{0.1, 0.2, 0.3, 0.4, 0.5\}$, $\gamma_{p} \in \{0.1, 0.3, 0.5, 0.7, 0.9\}$.

\section{Conclusion}
This paper proposes ENCOFA, a novel framework designed to address mixed open-set and closed-set label noise in medical image classification. 
The proposed ENCOFA leverages detected open-set noisy samples more effectively than existing methods. 
It incorporates an ENSC loss to enhance open-set sample identification and an OSFeatAug module to enrich open-set features, thereby reducing model overfitting to label noise. 
Our comparative experiments on two noisy datasets demonstrate the superior performance of the ENCOFA framework.

\noindent \textbf{Limitation} Although the ENCOFA framework is effective, its hyperparameter selection is sensitive to the noise level in the training data, leading to a substantial workload when adapting it to new datasets or tasks. 
Our future research is committed to mitigating this challenge by refining the adaptability and robustness of ENCOFA, to establish it as a dependable solution for managing label noise in a diverse range of medical image classification tasks.

\bibliographystyle{cas-model2-names}

\bibliography{cas-refs}

\end{document}